\documentclass{article}
\usepackage{spconf,amsmath,graphicx}

\title{End-to-End Facial Deep Learning Feature Compression with Teacher-Student Enhancement}
\name{Shurun Wang,   Wenhan Yang, Shiqi Wang}
\address{Department of Computer Science, City University of Hong Kong, Hong Kong, China}
\begin{document}
%
\maketitle
\begin{abstract}
In this paper, we propose a novel end-to-end feature compression scheme by leveraging the representation and learning capability of deep neural networks, towards intelligent front-end equipped analysis with promising accuracy and efficiency. 
In particular, the extracted features are compactly coded in an end-to-end manner by optimizing the rate-distortion cost to achieve feature-in-feature representation. In order to further improve the compression performance,  
we present a latent code level teacher-student  enhancement model, which could efficiently transfer the low bit-rate representation into a high bit rate one. Such a strategy further allows us to adaptively shift the representation cost to decoding computations, leading to more flexible feature compression with enhanced decoding capability. 
We verify the effectiveness of the proposed model with the facial feature, and experimental results reveal better compression performance in terms of rate-accuracy compared with existing models.
\end{abstract}
\begin{keywords}
Feature compression, deep learning, teacher-student network
\end{keywords}
\section{Introduction}
\label{sec:intro}
The deep learning features~\cite{lecun2015deep}, which are extracted with deep neural networks learned from abundant training data, have essential differences compared with handcrafted features, e.g., Histogram of Oriented Gradient (HOG) \cite{dalal2005histograms} and Scale-Invariant Feature (SIFT) \cite{lowe2004distinctive}. With the unprecedented success of deep learning in various computer vision tasks as well as  the development of network infrastructure, there is an increasing demand to study the deep learning feature compression in the  \textit{Analysis-then-Compress} (ATC)~\cite{6659301} paradigm. 
In particular, in contrast with \textit{Compress-then-Analysis} (CTA) paradigm where the videos would be first acquired at front-end sensors then compressed and transmitted to the cloud-end for analysis purposes, ATC allows the straightforward feature extraction at the front-end, leading to a much more compact representation of videos by transmitting the features instead of textures. In view of this advantage, the ATC paradigm with both handcrafted and deep learning features has been widely studied to address the challenges of video big data in various application scenarios. 



In the literature, there are numerous algorithms proposed for compact feature representation of both handcrafted and deep features. Hash-based model DBH \cite{6247912} and vector quantization based models, such as product quantization (PQ) \cite{5432202} optimized product quantization (OPQ) \cite{6678503}, target at the compact representation of handcrafted features. Moreover, binary based descriptors such as BRIEF \cite{6081878} and USB \cite{6832500} have been proposed for high-efficiency Hamming distance computation. Regarding deep learning features, 
Ding \textit{et al.} \cite{8016659} applied the philosophy of video coding to compact deep learning feature representation. The deep hashing network (DHN) \cite{zhu2016deep} combined supervised learning with hash compression to achieve performance promotion for image feature representation. Besides, Chen \textit{et al.} also proposed an intermediate deep feature compression towards intelligent sensing in \cite{8848858}.   


The promising characteristics of ATC paradigm motivate the standardization of the compact feature representation. In particular, the Compact Descriptor for Visual Search (CDVS) and Compact Descriptors for Video Analysis (CDVA) standards completed by the Moving Picture Experts Group (MPEG), define the standardized bitstream syntax such that the interoperability could be enabled in image/video retrieval applications. In 2019, the MPEG initiated the standardization of video coding for machine (VCM)~\cite{duan2020video,vcm-iso}, aiming to achieve high accuracy, low latency, object oriented analysis based on compact video representation for machine vision. 
VCM relies on the fundamental development of feature compression, and could establish the relationship between compact feature representation and video compression in terms of both machine vision and human perception, as features could be ultimately utilized in various machine vision tasks.  


\begin{figure*}[htbp]
\centerline{\includegraphics[width = 6.6in]{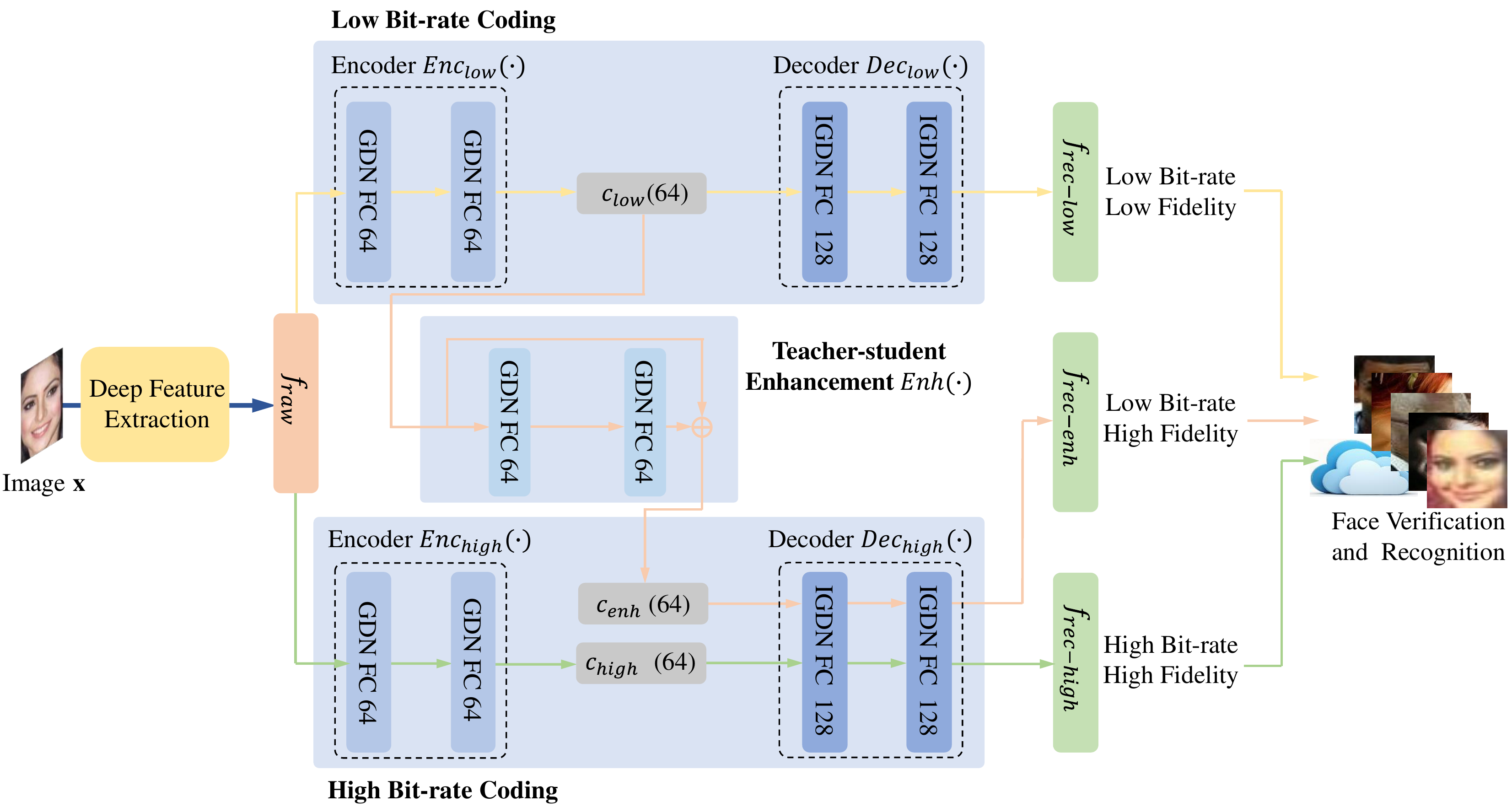}}
\vspace{-2mm}
\caption{The architecture of the proposed feature compression scheme.}
\label{framework}
\vspace{-4mm}
\end{figure*}

In this work, motivated by the recent progress on deep learning based video coding~\cite{8693636}, we attempt to further compress the raw deep learning features based on the representation and learning capability of deep neural works. The contributions of this paper are as follows,
\begin{itemize}
\item
We propose an end-to-end coding scheme to compactly represent the deep learning features as a latent code, in an effort to achieve optimal feature-in-feature representation based on the rate-distortion optimization. 

\item
We propose a compact feature enhancement method which further improves the feasibility in feature coding. The proposed scheme is built upon the teacher-student enhancement module at the latent code level, and allows to adaptively switch between high complexity decoding and high bit rate representation. 


\item
The proposed principled framework is implemented based on facial features, and better coding performance in terms of rate-accuracy has been demonstrated compared with the popular feature compression schemes. 
\end{itemize}



\section{The Framework of Feature Compression }
The architecture of the proposed scheme is shown in Fig. \ref{framework}. More specifically, the deep learning feature extraction from raw image $\textbf{x}$ with pre-trained FaceNet model\footnote{https://github.com/davidsandberg/facenet} is denoted as $f_{raw} = FaceNet(\textbf{x})$. Subsequently, the raw feature can be compressed with an end-to-end trained deep neural network, and for different bit rates different encoders and decoders are learned to adapt the characteristics of rate-distortion function. As such, the compact representations of $f_{raw}$ denoted as $c_{low}$ and $c_{high}$, indicate the compact latent code under low and high bit-rate scenarios, respectively. Moreover, the reconstructed features $f_{rec_{low}}$ and $f_{rec_{high}}$ can be obtained with $Dec_{low}$ and $Dec_{high}$, and this process can be formulated as follows,
\begin{equation}
    c_{low} = Enc_{low}(f_{raw}),~f_{rec_{low}}= Dec_{low}(c_{low}),
\end{equation}
\begin{equation}
    c_{high} = Enc_{high}(f_{raw}),~f_{rec_{high}} = Dec_{high}(c_{high}).
\end{equation}

Furthermore, the low bit rate stream \(c_{low}\) can be further enhanced by transferring it to the \(c_{high}\) as the target based on the teacher-student learning. As such, the output of the enhancement module can be well decoded with the decoder learned in the high bit-rate coding scenario. This process is expressed as follows,
\begin{equation}
    c_{enh} = Enh(c_{low}), \quad ~ f_{rec_{enh}} = Dec_{high}(c_{enh}).
\end{equation}


The reconstructed feature with the enhanced latent code $f_{rec_{enh}}$ reveals better fidelity compared with the reconstructed feature at low bitrate $f_{rec_{low}}$ by improving the representation capability with enhanced decoding process, as both enhancement and pure decoding should be performed sequentially in this scenario. In this manner, the flexibility of the feature codec is significantly improved in an effort to ensure the optimal rate-accuracy performance. 


\section{End-to-end feature compression with teacher-student enhancement}
\subsection{End-to-end feature compression}
\label{sec:pagestyle}
\begin{figure}[t]
\begin{minipage}[b]{1.0\linewidth}
  \centering
  \centerline{\includegraphics[width=8.9cm]{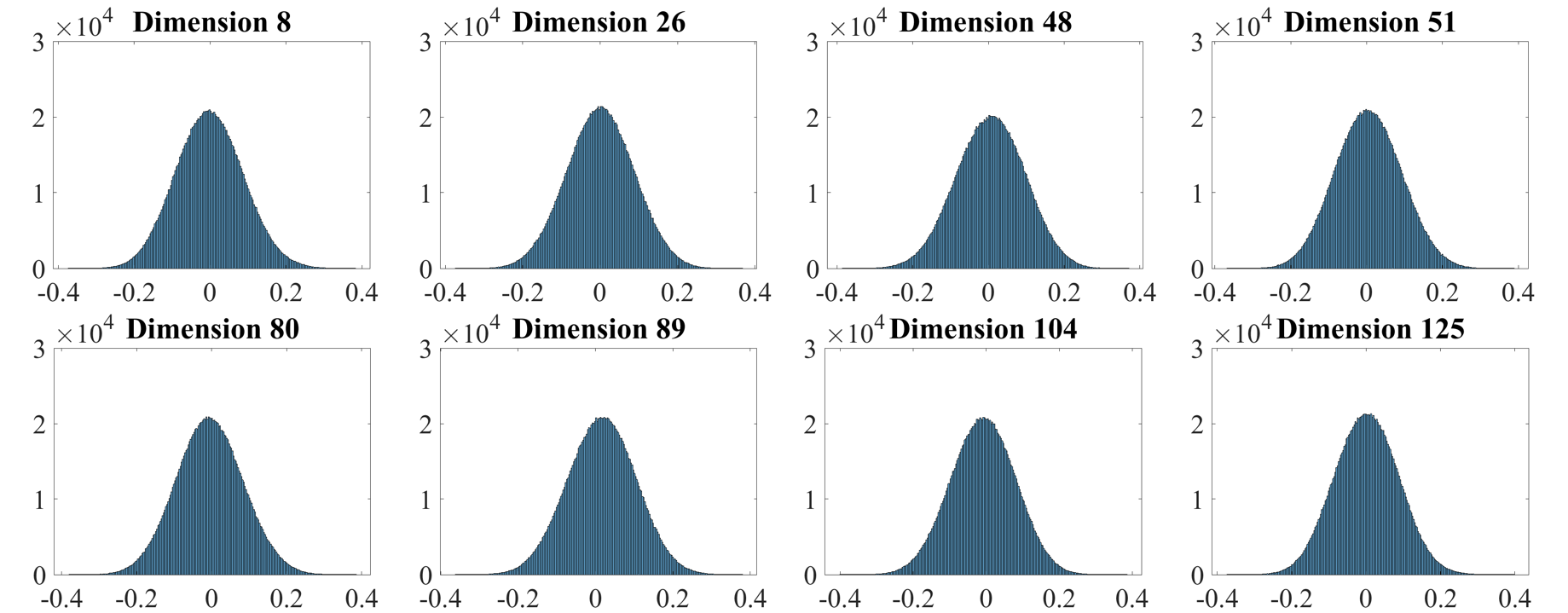}}
  \centerline{(a) VGG-Face2}\medskip
\end{minipage}
\hfill
\begin{minipage}[b]{1.0\linewidth}
  \centering
  \centerline{\includegraphics[width=8.9cm]{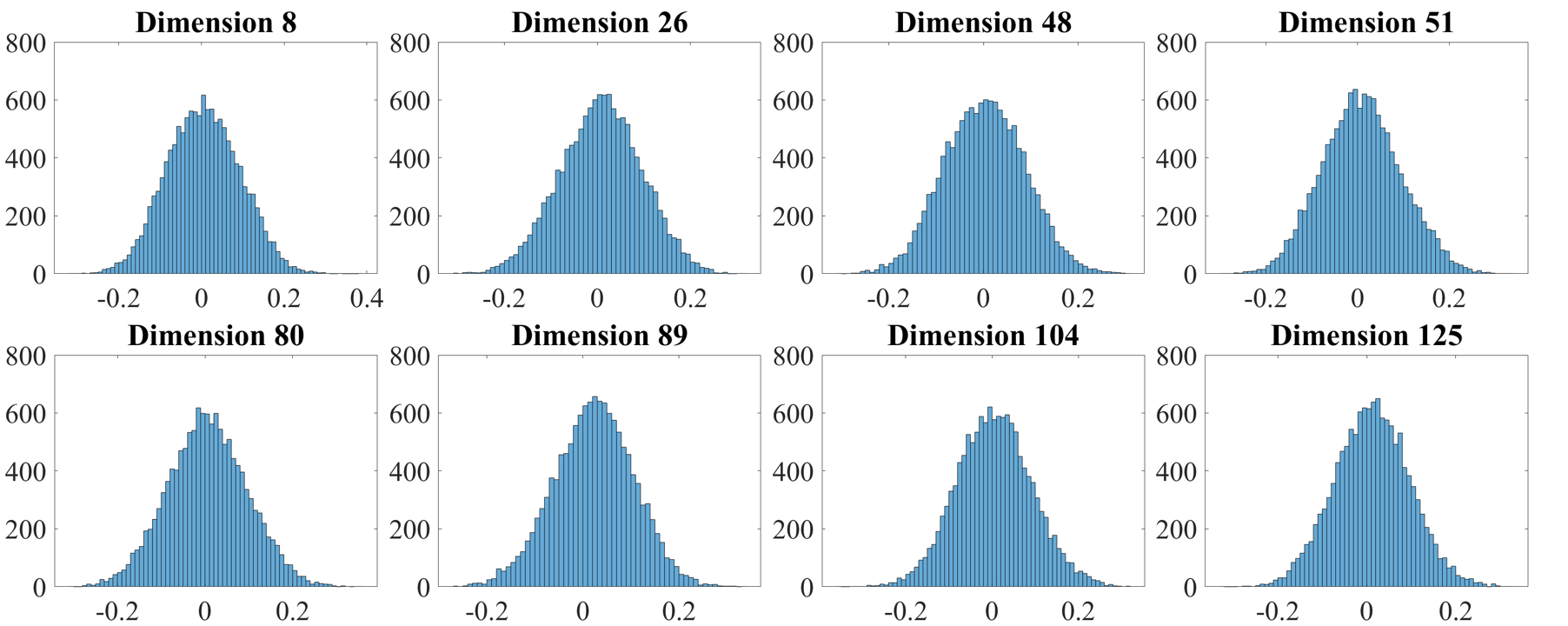}}
  \centerline{(b) LFW}\medskip
\end{minipage}
\vspace{-5mm}
\caption{The distribution of Facenet features extracted from several dimensions. (a) VGG-Face2; (b) LFW.}
\vspace{-3mm}
\label{distribution}
\end{figure}

To begin with, we extract deep learning features from the pre-trained FaceNet model, and investigate the distributions for end-to-end compression. The distribution of several dimensions in FaceNet features extracted from Labeled Face in Wild (LFW) \cite{LFWTech} and VGG-Face2 datasets \cite{Cao18} are shown in Fig.~\ref{distribution}. It is obvious to see the Gaussian-like distribution in the similar range and the expectations are all close to zero, indicating that the features well match the characteristics of generalized divisive normalization (GDN) in terms of Gaussianizing densities, as illustrated in \cite{7906310}. Motivated by the recent development of end-to-end image compression \cite{balle2018variational}, an end-to-end model by imposing $l_1$ norm as the sparsity constraint is trained for feature compression.

As illustrated in Fig. \ref{framework}, two fully-connected layers with GDN/IGDN \cite{balle2016end} are adopted as the Encoder/Decoder respectively, and an arithmetic coding engine is applied to generate the final bit-stream based on the latent code. The loss function includes the linear combination of mean square error (MSE) between original feature $f_{raw}$ and $f_{rec}$, and the $l_1$ norm value of compact representation denoted as $c$ to indicate the bit rate of the bitstream. The
balance between feature rate and distortion is governed by Lagrangian multiplier $\lambda$. The whole process is formulated as follows,
\begin{equation}\label{process1}
    c = Enc(f_{raw}),  \quad f_{rec} = Dec(c),
\end{equation}
\begin{equation}
    Loss_{coding} = ||f_{raw}-f_{rec}||_{2}^2+\lambda*||c||_{1}.
\end{equation}

Besides, the compact representation $c$ is clipped by the threshold $r_{clip}$ element-wise such that the expense in representing the feature is further reduced. It is worth mentioning that random noise is applied to simulate the distortion of the rounding operation for $c$ in the training process. 
Random noise also could strengthen the adaptation for the feature reconstruction in the decoder compared with quantization, which could be beneficial for the latent code level enhancement.

\subsection{Teacher-Student Enhancement at Latent Code level}
\label{sec:typestyle}
Based on the end-to-end feature compression, the teacher-student enhancement model is applied at the latent code level, to further improve the coding performance and feasibility.  
More specifically, the latent code for low bit rate coding $c_{low}$ is transferred to the high bit rate representation $c_{high}$, based on the correspondence between the two domains.
A straightforward approach is adopted here for teacher-student based enhancement, leading to the feasible solution that enhances the adaptively generated latent code with the specific domain knowledge based on a learned neural network. 


The structure of enhancement model is two fully-connected layers with GDN, as shown in Fig. ~\ref{framework}.  Range normalization is adopted as the data pre-processing by dividing the clipping threshold $r_{clip}$ used in the training of end-to-end feature compression.
As such, the loss function in learning the network that transfers $c_{low}$ to $c_{high}$ is defined as follows,
\begin{equation}
    Loss_{Enh} = ||c_{low}/r_{clip}-c_{high}/r_{clip}||_2^2.
\end{equation}


\subsection{Implementations}
TensorFlow \cite{abadi2016tensorflow} is adopted as the deep learning toolbox and the parameters in network are initialized with the method in  \cite{glorot2010understanding}. The batch size is set to 32. Moreover, the learning rate is set to 0.0001 for the stable convergence during training and the epoch of training is 40. Adaptive Moment Estimation (Adam) \cite{kingma2014adam} is adopted as the optimizer algorithm in all deep learning models. Regarding the end-to-end feature compression model, $\lambda$ is set from $1\times10^{-4}$ to $1\times10^{-7}$ to learn different models from low bit-rate to high bit-rate coding. The threshold for the latent code value clipping is set to 20.0 and the random noise range is set from $-0.5$ to $0.5$. 

\section{Experimental Results}
\label{sec:majhead}


\begin{figure}[t]
\centerline{\includegraphics[width = 2.7in]{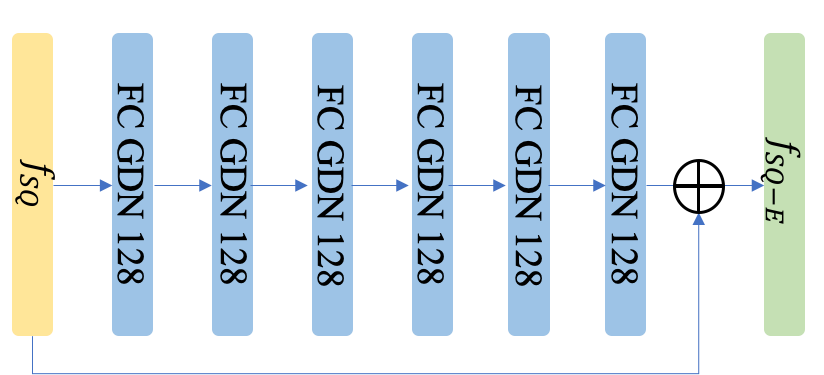}}
\caption{Illustration of the model structure of SQ-E.}
\label{aug_gdn}
\end{figure}

\begin{table*}[htbp]
\footnotesize
\caption{ Compression performance comparison in terms of rate-accuracy with various models.}
\begin{tabular}{|cc|cc|cc|cc|cc|cc|cc|cc|}
\hline
\multicolumn{2}{|c|}{PQ} & \multicolumn{2}{c|}{OPQ} & \multicolumn{2}{c|}{DBH} & \multicolumn{2}{c|}{DHN} & \multicolumn{2}{c|}{SQ} & \multicolumn{2}{c|}{SQ-E} & \multicolumn{2}{c|}{PRO} & \multicolumn{2}{c|}{PRO-E}      \\ \hline
BPP     & Acc(\%)   & BPP     & Acc(\%)   & BPP     & Acc(\%)   & BPP     & Acc(\%)   & BPP     & Acc(\%)   & BPP      & Acc(\%)    & BPP     & Acc(\%)   & BPP           & Acc(\%)    \\ \hline
1.00    & 98.48          & 1.00    & 98.40          & 1.00    & 97.48          & 1.00    & 98.25          & 1.21    & 50.00          & 1.21     & 50.41           & 0.81    & 98.28         & \textbf{0.81} & \textbf{98.63} \\ 
2.00    & 99.13          & 2.00    & 99.17          & 2.00    & 98.23          & 2.00    & 98.70          & 1.70    & 98.60          & 1.70     & 98.72           & 1.42    & 98.96         & \textbf{1.42} & \textbf{99.10}  \\ 
4.00    & 99.28          & 4.00    & 99.25          & 4.00    & 98.43          & 4.00    & 99.08          & 2.58    & 99.11          & 2.58     & 99.15           & 2.03    & 99.27          & \textbf{2.03} & \textbf{99.30}  \\ 
8.00    & 99.25          & 8.00    & 99.27          & 8.00    & 98.83          & 8.00    & 99.13          & 4.21    & 99.23          & 4.21     & 99.26           & 2.61    & 99.27          & \textbf{2.61} & \textbf{99.30}  \\ \hline
\end{tabular}
\end{table*}

We conduct experiments to validate the effectiveness of the proposed models in terms of rate-accuracy. The training data are VGG-Face2 \cite{Cao18} with over 3.3 million human face images, including 9131 subjects and every subject has over 360 images on average. Correspondingly, the testing data are popular face verification dataset, Labeled Faces in the Wild (LFW) \cite{LFWTech}. 

In order to verify the effectiveness of the proposed models, we adopt the scalar quantization algorithm (SQ) used in \cite{8803255} for comparison. Moreover, on top of this strategy, we introduce a deep learning based feature enhancement model (SQ-E) for further comparisons. In particular, FaceNet feature is 128-dimension vector in range of -1 and 1 and the SQ based compression is conducted with the following procedures,
\begin{equation}
q_{step} = 2^{\frac{QP-4}{6}-10},
\end{equation}
\begin{equation}
    c_{SQ} = floor(f_{raw}/q_{step}),
\end{equation}
\begin{equation}
    f_{SQ} = c_{SQ}*q_{step}.
\end{equation}
where $f_{SQ}$ and $f_{raw}$ are the reconstructed and original feature respectively, and $QP$ is quantization parameter. $c_{SQ}$ is the quantized feature, which is further subjected to entropy coding to generate the feature bitstream. 

\begin{figure}[t]
	\centering
	\centerline{\includegraphics[width=7.1cm]{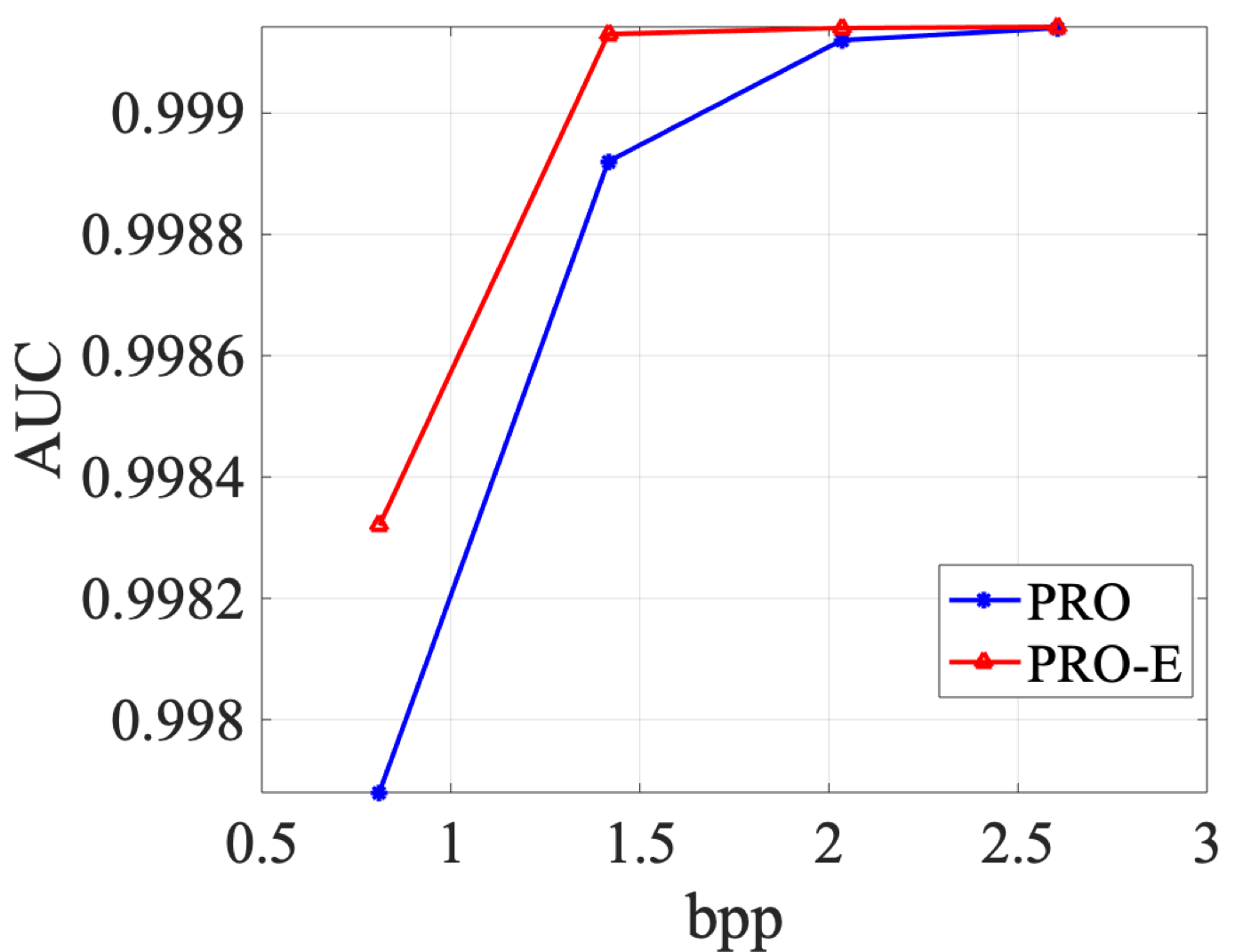}}
	\caption{Compression performance comparisons in terms of  Bitrate-AUC.}
	\label{curves}
	\vspace{-3mm}
\end{figure}

\begin{figure}[t]
	\centering
	\centerline{\includegraphics[width=7.1cm]{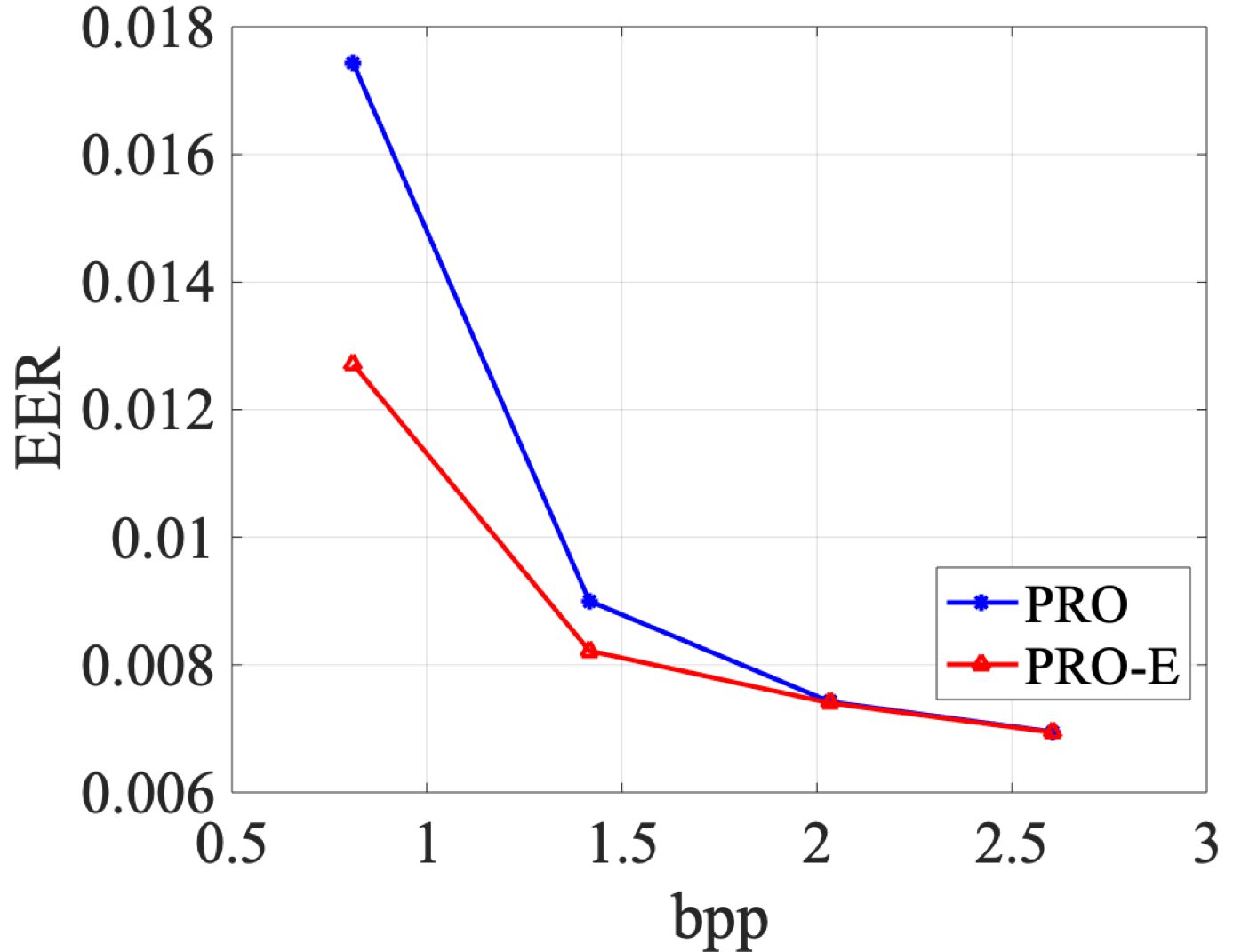}}
	\caption{Compression performance comparisons  in terms of Bitrate-EER.}
	\label{curves2}
	\vspace{-3mm}	
\end{figure}

Moreover, in order to validate the proposed scheme with a comparable deep learning model as the baseline, a neural network based model for SQ reconstructed feature is introduced to enhance the feature fidelity via the residual network and GDN, denoted as SQ-E. The network structure is shown in Fig.~\ref{aug_gdn}. The loss function of the network is mean square error (MSE) between original feature $f_{raw}$ and enhanced feature $f_{SQ-E}$ to enhance the quality of the decompressed feature.

We first verify the effectiveness of the proposed scheme in terms of rate-accuracy performance, as shown in Table 1. In particular, the proposed end-to-end feature compression model and the teacher-student enhancement model are denoted as PRO and PRO-E, respectively. It is also worth mentioning that accuracy of the original FaceNet feature without compression is 99.32\% with a public pre-trained FaceNet model. In addition to SQ and SQ-E, other compression algorithms including PQ \cite{5432202}, OPQ \cite{6678503}, DBH \cite{6247912}, DHN \cite{zhu2016deep} are also compared \cite{avs}. It is obvious that the proposed scheme could achieve better compression performance in terms of the rate-accuracy. Moreover, in order to investigate the performance of the teacher-student enhancement model, the area under curve (AUC) and equal error rate (EER) performance between PRO and PRO-E are also compared, as shown in Fig.~\ref{curves} and~\ref{curves2}. The rate-accuracy curves provide useful evidence regarding the effectiveness of the proposed enhancement model. 





\section{Conclusions}
\label{sec:print}

In this paper, we propose an end-to-end deep feature coding framework towards video coding for machine. 
The novelty of this paper lies in that  instead of directly quantizing and entropy coding the features, we introduce the deep learning model to further compactly represent the features as the latent code,  such that better performance can be achieved. Moreover, a feature enhancement approach is proposed at the latent code level, which transfers the low quality latent code representation into a high quality one to facilitate the subsequent analysis process. 
Experiments have proven the efficiency of the proposed deep learning feature representation scheme from different perspectives. 


\renewcommand{\baselinestretch}{0.85}


\end{document}